\documentclass[10pt,letterpaper]{article}
\usepackage[T1]{fontenc}
\usepackage[utf8]{inputenc}
\usepackage{preprint,times}

\usepackage{amsmath,amsfonts,bm}

\def\eqref#1{equation~\ref{#1}}

\def\1{\bm{1}}

\DeclareMathAlphabet{\mathsfit}{\encodingdefault}{\sfdefault}{m}{sl}
\SetMathAlphabet{\mathsfit}{bold}{\encodingdefault}{\sfdefault}{bx}{n}

\usepackage{graphicx}
\usepackage[table]{xcolor}
\usepackage{booktabs,multirow,array}
\usepackage{float,flafter}
\usepackage{microtype}
\usepackage{url}
\usepackage[colorlinks=true,allcolors=blue]{hyperref}
\hypersetup{
  pdftitle={Real2SAM2Real: Generative 3D Caches as Complementary Context for Video Diffusion},
  pdfauthor={},
  pdfsubject={3D-aware controllable video generation},
  pdfkeywords={Video diffusion, 3D-aware video generation, Camera and motion control},
  pdfcreator={LaTeX}
}


\title{Real2SAM2Real: Generative 3D Caches as Complementary Context for Video Diffusion}
\author{
  \normalsize Jiayi Wu\thanks{Equal contribution.} \quad Haoming Cai\footnotemark[1] \quad Cornelia Fermuller\\
  \normalsize Christopher Metzler \quad Yiannis Aloimonos\\[0.5em]
  \small University of Maryland, College Park\\[17pt]
  {\large\bfseries\urlstyle{same}Project page: \url{https://jiayi-wu-leo.github.io/real2sam2real/}}
}
\hypersetup{
  pdfauthor={Jiayi Wu, Haoming Cai, Cornelia Fermuller, Christopher Metzler, Yiannis Aloimonos}
}

\date{}

\begin{document}
\pagestyle{plain}
\raggedbottom
\maketitle
\begin{abstract}

Video Diffusion Models (VDMs) synthesize high-quality videos, but precise camera and motion control remains challenging when large viewpoint shifts, changing occlusions, or complex dynamics reveal unseen regions. Diffusion priors alone can yield structural inconsistencies, while warped guidance can contain incomplete geometry and misleading view-dependent appearance cues. Real2SAM2Real complements VDM priors with a coarse 3D foreground cache of topologically complete meshes for intuitive editing and animation. Generative 3D lifting estimates these object and human meshes, including unobserved surfaces, from a single reference image. Normal maps rendered under specified camera trajectories and foreground motions guide video synthesis in the VDM, while the reference image provides appearance conditioning. Soft spatial-aligned injection and minimally invasive fine-tuning incorporate this guidance while largely preserving pretrained synthesis priors. Decoupled conditioning enables independent appearance and geometry editing. We train on around 2,000 monocular clips using instance-masked normal sequences with instance-level spatial perturbations to reduce sensitivity to differences between video-estimated training normals and normals rendered from coarse 3D caches at inference. Training requires neither explicit 3D assets nor paired multi-trajectory recordings. Real2SAM2Real supports decoupled camera and multi-instance motion control, including articulated humans. Across these challenging settings, experiments show strong spatiotemporal consistency and better overall control accuracy and video fidelity than evaluated state-of-the-art methods. Preliminary demonstrations suggest potential robotics applications.
\end{abstract}

\begin{figure}[!t]
    \centering
    \includegraphics[width=0.99\linewidth]{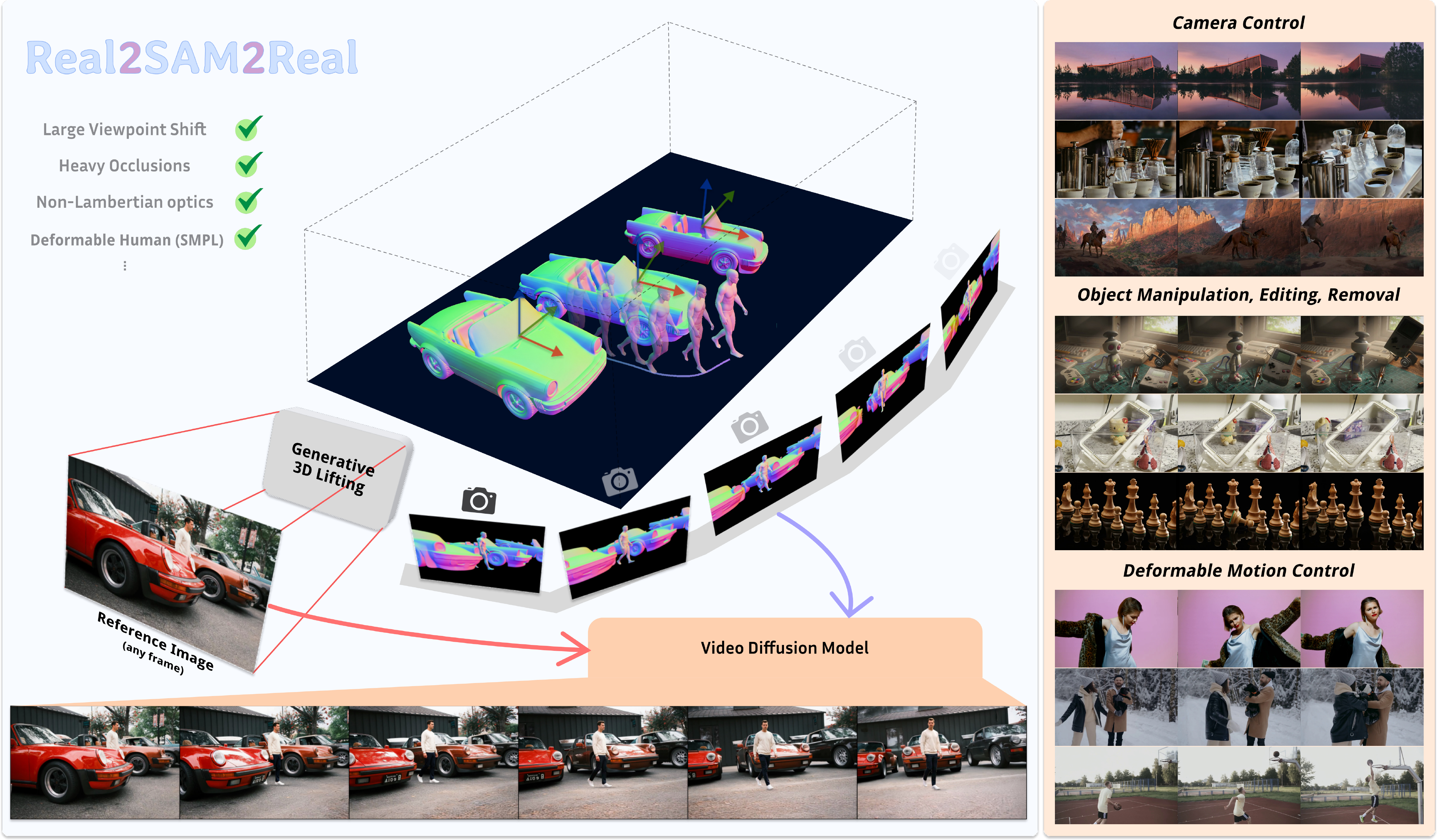}
    \caption{\textbf{Versatile 3D-Aware Video Generation via Real2SAM2Real.} \textbf{(Left)} From a single reference image at any frame, generative 3D lifting constructs an editable, \textit{instance-complete} cache of objects and SMPL-based humans. The VDM uses coarse foreground geometry rendered according to the specified camera trajectory and foreground motion as guidance. \textbf{(Right)} Applications include camera control, entity manipulation and removal, and deformable motion control, with examples involving occlusions, viewpoint ambiguity, human motion, reflections, refractions, and flames.
    }
    \label{fig:teasor}
\end{figure}
\section{Introduction}
Video Diffusion Models (VDMs)~\citep{wan2025wan,yang2024cogvideox,hacohen2024ltx,kong2024hunyuanvideo,blattmann2023stable} synthesize high-fidelity videos using spatiotemporal priors learned from large-scale datasets. Recent work adapts these models through post-training~\citep{zhang2023adding,hu2022lora,mou2024t2i} for fine-grained camera and entity motion control. However, their training observations are 2D videos. While an ideal world model should simulate complete 4D spacetime, videos intrinsically capture dynamics solely on a projected 2D plane. This dimensional loss can limit geometric fidelity, spatiotemporal consistency, and explicit 3D-aware control in complex dynamic scenes.

Existing methods largely follow two paradigms. The first fine-tunes VDMs with parameterized camera embeddings~\citep{he2024cameractrl,wang2024motionctrl,bai2025recammaster, van2024generative,luo2025camclonemaster,bai2024syncammaster}. This can require additional adaptation and risk degrading pretrained priors, while scale ambiguity may remain. The second uses pixel-wise 3D reconstructions, such as point clouds, in a "warp-and-inpaint" pipeline~\citep{yu2024viewcrafter,yu2025trajectorycrafter,ren2025gen3c,song2025worldforge,you2024nvs,liu2024novel}. Although this reduces the modality gap, incomplete geometry and misleading cues in reflective or refractive regions can produce perspective ambiguities and artifacts. When motion exposes unobserved regions, insufficient geometric guidance can make structural consistency difficult to maintain.

We introduce \textbf{Real2SAM2Real}, which complements VDM priors with an editable, instance-complete foreground geometry cache (Fig.~\ref{fig:teasor}). From a single reference image, generative 3D lifting~\citep{chen2025sam,xiang2025structured,hunyuan3d2025hunyuan3d,wu2025direct3d,ye2025hi3dgen} builds a coarse cache that provides geometric priors and an intuitive interface for camera control and editing scene and character layouts. Background synthesis is left to the VDM. Coarse foreground geometry provides scene-anchored visual guidance, while minimally invasive fine-tuning aims to retain pretrained capabilities. Experiments show geometrically coherent generation with precise control in evaluated scenes involving large motions, heavy occlusions, and non-Lambertian effects (Fig.~\ref{fig:Adv}).

The core contributions of this work can be summarized as follows:
\begin{enumerate}
    \setlength{\itemsep}{2pt}
    \setlength{\parsep}{0pt}
    \setlength{\parskip}{0pt}
    \item \textbf{We introduce a topologically complete 3D geometry cache for intuitive editing and animation.} These spatial anchors support camera and entity control while helping mitigate artifacts from incomplete geometry (e.g., holes and erroneous facades) and misleading cues in reflective or refractive regions.
    \item \textbf{We use instance-masked normal sequences as a cross-modal bridge between the 3D cache and the VDM.} Normal maps provide dense, scale-invariant geometry without texture details, complementing the appearance condition. We derive RGB-normal training pairs directly from monocular videos, without paired multi-trajectory recordings or 3D assets, and apply instance-level random perturbations to the normal maps to help reduce the training-inference gap.
    \item \textbf{We design a Soft Spatial-Aligned Injection mechanism and a minimally invasive fine-tuning strategy to decouple appearance and geometry, enabling independent editing of both.} This design adapts the VDM by training approximately 1.8\% of its parameters on around 2,000 video clips, while spatial perturbations help accommodate imperfect geometric guidance.
\end{enumerate}

\clearpage
\section{Background \& Related Works}

\begin{figure}[!htbp]
    \centering
    \includegraphics[width=0.99\linewidth]{figures/Adv.pdf}
    \caption{\textbf{Addressing the limitations of warp-and-inpaint pipelines.} Point-cloud warping can entangle geometry with view-dependent appearance: (a) reflections and (b) refractions introduce misleading spatial cues. (c) Incomplete geometry, background depth errors, and the geometric encoding of unstructured dynamics (e.g., fire) can produce stretched textures and erroneous facades under large viewpoint shifts. Our appearance-decoupled, \textit{instance-complete} cache reduces these artifacts while maintaining spatiotemporal coherence.}
    \label{fig:Adv}
\end{figure}

\subsection{Camera Control}
Current methodologies for camera control can be broadly classified into two dominant paradigms. The first paradigm attempts to adapt foundational video models to parameterized camera embeddings. 
Methods~\citep{he2024cameractrl,wang2024motionctrl,bai2025recammaster, van2024generative,luo2025camclonemaster,bai2024syncammaster}
directly inject camera trajectories (e.g., $[\mathbf{R}|\mathbf{t}]$ extrinsic matrices or Plücker rays) into the generative backbone through various attention mechanisms, feature concatenation, or similar conditioning modules. While integrating camera-centric parameterized control signals enables explicit trajectory controllability, this approach is limited by scale ambiguity and a cross-modal gap. Aligning these signals with visual features requires additional adaptation.

Instead of numerical poses, the second paradigm relies on reprojecting explicit pixel-wise 3D reconstructions as dense spatial guidance. Methods~\citep{yu2024viewcrafter,yu2025trajectorycrafter,ren2025gen3c,song2025worldforge,hu2025ex,ma2025you,you2024nvs,xiao2024trajectory,cao2025uni3c,liu2024novel,zhang2025recapture}
lift monocular depth into 3D point clouds or tracks~\citep{wang2025vggt,piccinelli2024unidepth,piccinelli2025unidepthv2,wang2025moge,bochkovskii2024depth,depthanything3,huang2025vipe}, warp them to novel views, and condition the VDM on these proxy videos. While this "warp-and-inpaint" strategy mitigates the cross-modal gap, the imperfections of monocular reconstructions expose the model to geometric errors. Specifically, the unclosed nature of these visible shells can introduce perspective ambiguities (via structural holes and erroneous facades), while non-Lambertian regions (e.g., reflections and refractions) yield misleading geometric cues. Together, these flawed spatial proxies can trigger cascading artifacts and structural degradation. Both paradigms provide limited explicit 3D context for unobserved regions. When large camera movements unveil previously hidden surfaces, this limited geometric guidance leaves the model to infer missing geometry from its generative priors, which can lead to structural degradation.
\subsection{Motion Control}
While camera control defines the global observation window, motion control guides the motion and interactions of objects and articulated subjects. This domain exhibits significant diversity, predominantly relying on a wide spectrum of abstract object representations or pixel-level cues as conditioning signals. For instance, 3DTrajMaster \citep{fu20243dtrajmaster} explicitly conditions the generation on the numerical embeddings of 6-DoF object pose trajectories. For articulated motion, Motion-2-to-3 \citep{pi2024motion} uses 2D motion data for 3D human motion generation. Motion Prompting \citep{geng2025motion} uses point trajectories, while Diffusion as Shader \citep{gu2025diffusion} conditions generation on 3D tracking videos. Uni3C \citep{cao2025uni3c} combines SMPL-X human meshes with monocular scene point clouds. For non-human foreground objects, these conditions provide limited geometric coverage, leaving the VDM to infer unobserved surfaces from its spatiotemporal priors. This reliance can lead to structural degradation under large motions and complex occlusions.

To address these limitations, our \textbf{Real2SAM2Real} framework introduces an explicitly editable, \textit{instance-complete} 3D proxy for objects and SMPL-based humans via instance-masked normal maps. This appearance-decoupled geometric guidance provides priors for unobserved regions, mitigating the ``inpainting trap" and cross-modal alignment issues of prior methods. Furthermore, by integrating this 3D context in a minimally invasive and fault-tolerant manner, our approach largely preserves foundational VDM priors while enabling 3D control and easy editing of diverse foreground instances.

\section{Methodology}

\subsection{Entities as Anchors: Instance-complete 3D Geometry Cache}
To synthesize a spatiotemporally consistent video $V \in \mathbb{R}^{T \times H \times W \times 3}$ of length $T$, we take as input a single reference image $I_{ref} \in \mathbb{R}^{H \times W \times 3}$, a user-specified camera trajectory $\mathbf{C} = \{c_t\}_{t=1}^T$, and a set of instance-specific spatial manipulation sequences $\mathbf{M} = \{\mathbf{M}_k\}_{k=1}^K$. We decouple spatial layout from appearance through an editable 3D staging environment. Full-scene reconstruction from a single image is ambiguous, particularly for distant backgrounds. We therefore model foreground instances explicitly and use the VDM's pretrained priors for background synthesis. Let $\Phi_{\text{3D}}$ denote a layout-aware, multi-instance 3D generation framework (e.g., SAM3D~\citep{chen2025sam}). The initial static 3D representation is formulated as the union of decoupled 3D foreground entities:
\begin{equation}
    \mathcal{S}_{\text{3D}}^{(0)} = \Phi_{\text{3D}}(I_{ref}) = \bigcup_{k=1}^K O_k
\end{equation}
where $O_k$ denotes the 3D mesh of the $k$-th instance, including objects and SMPL-based humans. To enable intuitive motion control, we apply the user-specified temporal editing parameters $m_{k,t} \in \mathbf{M}_k$ (e.g., 6-DoF rigid transformations or non-rigid deformations) to each decoupled instance. The dynamic scene configuration at any temporal frame $t$ is formulated as:
\begin{equation}
    \mathcal{S}_{\text{3D}}^{(t)} = \bigcup_{k=1}^K \mathcal{T}(O_k, m_{k,t})
\end{equation}
where $\mathcal{T}(\cdot)$ denotes geometric manipulation. Here, topological completeness refers to full instance surfaces, including estimated unobserved regions. These independently animatable meshes determine depth ordering and occlusions during rendering, reducing structural holes and cardboard effects caused by incomplete proxies.

\subsection{Normal as A Bridge: 3D-Free Data Curation \& Random Perturbation}
\label{sec3:data_curation}
To translate these coupled explicit 3D priors (both camera and entity dynamics) into a visual domain easily digestible by the VDM, we employ a graphics renderer $\mathcal{R}$. The dynamic 3D cache $\mathcal{S}_{\text{3D}}^{(t)}$ is projected along the target camera pose $c_t$ to yield a sequence of instance-masked normal maps $\mathbf{N} \in \mathbb{R}^{T \times H \times W \times 3}$:
\begin{equation}
    \mathbf{N} = \{n_t\}_{t=1}^T, \quad \text{where } n_t = \mathcal{R}(\mathcal{S}_{\text{3D}}^{(t)}, c_t)
\end{equation}
Normal maps encode local surface orientations independently of scene scale, and their three-channel format is compatible with the pretrained VAE. We construct RGB-normal pairs directly from monocular videos, without paired multi-trajectory recordings, explicit 3D assets, or reprojected conditioning videos. NormalCrafter~\citep{bin2025normalcrafter} estimates pseudo-normal sequences; SAM 3~\citep{carion2025sam} and vision-language models identify foreground instances, whose masks are applied to the estimated normals to obtain $\mathbf{N}$. To reduce the training-inference gap caused by noisy geometry and coarse boundaries in rendered 3D caches, we randomly perturb each instance's normal sequence with translation, rotation, non-uniform scaling, and boundary displacement. Each instance shares one perturbation parameter set across all frames, while RGB targets remain unchanged. This reduces sensitivity to imperfect geometric guidance.

\subsection{VDM as A Neural Compositor: Soft Spatial-Aligned Injection and Minimally Invasive Fine-Tuning}

\begin{figure}[!htbp]
    \centering
    \includegraphics[width=0.99\linewidth]{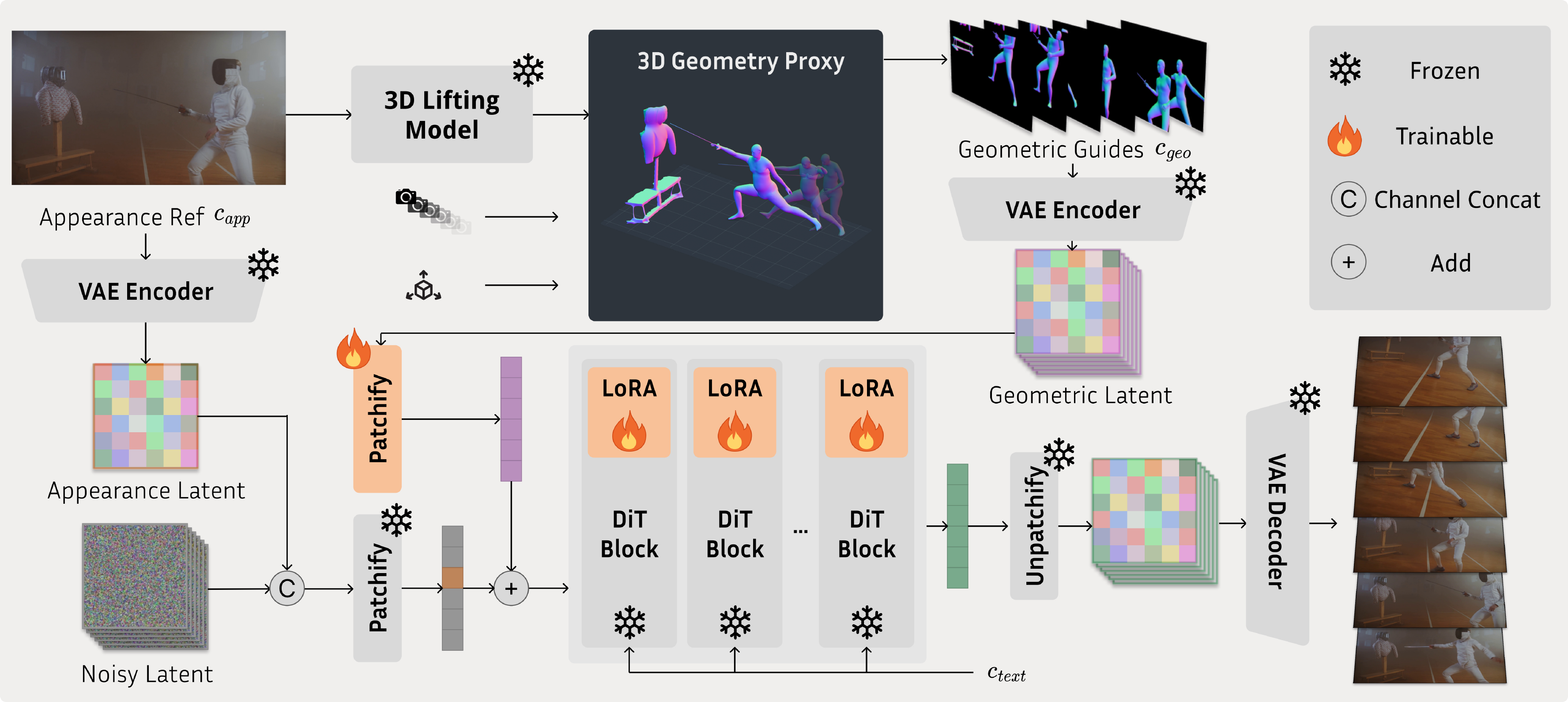}
    \caption{\textbf{Pipeline Architecture and Decoupled Injection Mechanism.} The reference frame enters through the backbone's native inpainting condition. A frozen VAE encodes instance-masked normal sequences, and a zero-initialized 3D convolution maps these latents to additive video features. A frozen T5 encoder supplies optional text conditioning. For each noise expert, only the geometry embedding layer and LoRA modules are trained.}
    \label{fig:train_pipeline}
\end{figure}
Our generative backbone is Wan2.2-Fun-A14B-InP, with high- and low-noise DiT experts. Let $z_0=\mathcal{E}(V)$ denote the video latent and $z_s=(1-\sigma_s)z_0+\sigma_s\epsilon$, where $\sigma_s$ is the noise level at timestep $s$ and $\epsilon\sim\mathcal{N}(0,I)$. We train a conditional velocity predictor $v_\theta$ using flow matching:
\begin{equation}
    \mathcal{L}=\mathbb{E}_{z_0,\epsilon,s}\left[\rho\!\left(v_\theta(z_s,s,\mathbf{c}_{app},\mathbf{c}_{geo},\mathbf{c}_{text})-(\epsilon-z_0)\right)\right],
\end{equation}
where $\rho$ averages squared residuals with outlier masking. The appearance condition $\mathbf{c}_{app}$ combines a binary mask with VAE latents of a video containing the reference image at its designated temporal position. It enters the backbone's native inpainting pathway through channel concatenation with $z_s$. During training, the reference frame is sampled from the target video. Optional text guidance $\mathbf{c}_{text}$ is encoded by a frozen T5 encoder.

For geometric guidance, the frozen VAE encodes $\mathbf{N}$, and a zero-initialized, trainable 3D convolution with a $1\times2\times2$ kernel maps these latents to geometric features $\mathbf{c}_{geo}$. These features are added to the frozen input patch embedding's output across all video frames. This \textit{soft spatial-aligned injection} uses a separate additive branch for geometry while retaining the pretrained inpainting pathway for appearance. The two conditions can be edited independently; motion transfer is performed by editing the reference image while keeping the normal-map sequence fixed.

We use \textit{minimally invasive fine-tuning}, training only the geometry patch embedding and LoRA modules in attention and feed-forward projections, totaling approximately 1.8\% of the base parameters per expert. The two noise experts are fine-tuned separately on their corresponding noise ranges, while the pretrained DiT weights, VAE, and T5 encoder remain frozen. Combined with instance-level normal perturbations, these limited updates reduce overfitting to imperfect geometry while largely preserving the backbone's appearance and background synthesis priors. The native text-conditioning pathway supports optional editing of visual effects and styles alongside 3D control.

\section{Experimental Results}

\subsection{Datasets and Experimental Setup}

We implement our framework based on Wan2.2-Fun-A14B-InP and fine-tune it using around 2,000 monocular video clips, from which our 3D-free pipeline automatically constructs training pairs of RGB videos and randomly perturbed instance-masked normal sequences. We fine-tune the high- and low-noise experts separately at 720p (81 frames), using LoRA and trainable geometry embeddings.
Details of the training data curation and training setup are provided in Appendix~\ref{supp_impl}, and the test data, baseline conditioning, and evaluation metrics of each experiment in Appendix~\ref{supp_exp}.

\subsection{Camera Control}

We evaluate on 200 test cases constructed from 100 in-the-wild images (Unsplash), each paired with 2 randomly sampled GLB camera trajectories, with ground-truth poses directly recorded from the GLB environment.
We compare against ReCamMaster~\citep{bai2025recammaster}, ViewCrafter~\citep{yu2024viewcrafter}, TrajectoryCrafter~\citep{yu2025trajectorycrafter},
WorldForge~\citep{song2025worldforge}, and GEN3C~\citep{ren2025gen3c}. Warp-based baselines are conditioned on an identical WorldForge-generated proxy for fairness, and GEN3C renders its own 3D cache along the same trajectory; ReCamMaster receives ground-truth poses converted to its required format. Camera fidelity is measured by ATE, RPE-T, and RPE-R via GLOMAP~\citep{pan2024global} and evo~\citep{grupp2017evo}, following~\citep{song2025worldforge,bai2025recammaster}, with VBench~\citep{huang2024vbench} for
video quality. As shown in Tab.~\ref{tab:camera_control_quantitative} and Fig.~\ref{fig:main_03_exp_cam_control},
Baseline methods exhibit cascading artifacts under large viewpoint shifts, whereas Real2SAM2Real maintains faithful appearance and spatiotemporal coherence in the evaluated scenes.

\begingroup
\setlength{\intextsep}{4pt plus 1pt minus 1pt}
\providecommand{\first}[1]{\cellcolor{red!12}\textbf{#1}}
\providecommand{\second}[1]{\cellcolor{blue!12}#1}
\begin{table}[!htbp]
\centering
\setlength{\tabcolsep}{3.5pt}
\setlength{\aboverulesep}{0pt}\setlength{\belowrulesep}{0pt}\renewcommand{\arraystretch}{1.05}
\setlength{\abovecaptionskip}{0pt}\setlength{\belowcaptionskip}{3pt}
\caption{Quantitative comparison of camera control on 200 test cases.
Camera accuracy is reported by ATE, RPE-T and RPE-R, computed with GLOMAP~\citep{pan2024global} and evo~\citep{grupp2017evo} following WorldForge~\citep{song2025worldforge} and ReCamMaster~\citep{bai2025recammaster}; video quality is reported by VBench~\citep{huang2024vbench}.
The best and second-best results are shaded in \colorbox{red!12}{red} and \colorbox{blue!12}{blue}; ablations are not ranked.}
\label{tab:camera_control_quantitative}
\resizebox{\textwidth}{!}{%
\begin{tabular}{l cccc ccc}
\toprule
\multirow{2}{*}{\textbf{Method}} & \multicolumn{4}{c}{\textbf{VBench} ($\uparrow$)} & \multicolumn{3}{c}{\textbf{Camera Accuracy} ($\downarrow$)} \\
\cmidrule(lr){2-5} \cmidrule(lr){6-8}
 & Subject & Background & Aesthetic & Imaging & ATE & RPE-T & RPE-R \\
\midrule
ReCamMaster~\citep{bai2025recammaster} & 90.29 & 91.39 & 55.54 & 68.98 & 0.0620 & 0.0182 & \first{1.0649} \\
ViewCrafter~\citep{yu2024viewcrafter} & 89.95 & 91.34 & \second{62.03} & \second{74.20} & 0.0809 & 0.0712 & 3.6352 \\
TrajectoryCrafter~\citep{yu2025trajectorycrafter} & 90.63 & 93.01 & 57.42 & 74.02 & 0.0443 & 0.0286 & 1.8461 \\
WorldForge~\citep{song2025worldforge} & 89.92 & 92.73 & 56.87 & 70.87 & 0.1281 & 0.0605 & 2.2211 \\
GEN3C~\citep{ren2025gen3c} & \first{91.42} & \second{93.32} & 57.85 & 73.13 & \first{0.0210} & \second{0.0143} & 1.1461 \\
\midrule
Ours (w/o Soft Injection) & 86.26 & 89.80 & 59.60 & 71.67 & 0.0553 & 0.0268 & 1.1309 \\
Ours (w/o Perturbation) & 89.19 & 91.75 & 61.40 & 75.04 & 0.0280 & 0.0148 & 1.1553 \\
Ours (w/o Both) & 86.22 & 89.76 & 59.58 & 71.69 & 0.0645 & 0.0265 & 1.2557 \\
\midrule
\textbf{Ours (Full)} & \second{91.33} & \first{93.70} & \first{62.12} & \first{75.09} & \second{0.0234} & \first{0.0132} & \second{1.0940} \\
\bottomrule
\end{tabular}%
}
\end{table}

\begin{figure}[!htbp]
    \centering
    \includegraphics[width=0.99\linewidth]{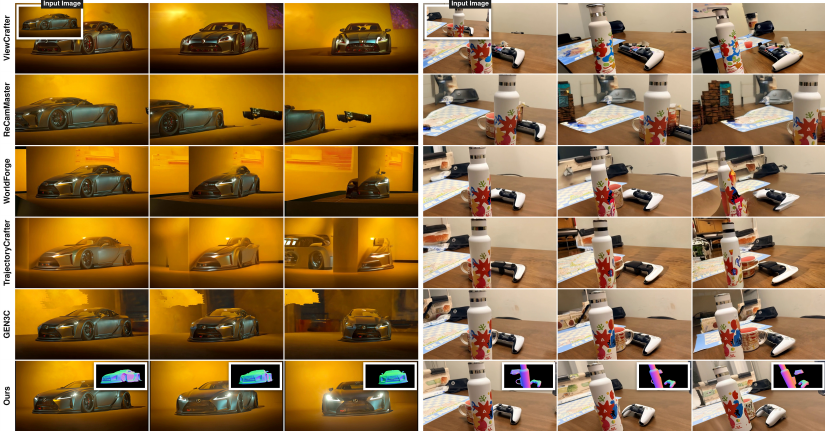}
    \caption{
    \textbf{Qualitative comparison of camera control.}
    Two scenes are shown side by side; within each, frames advance in time from left to right. Baseline methods exhibit cascading artifacts and structural breakdowns under large viewpoint shifts. Ours maintains faithful appearance and spatiotemporal coherence in the shown scenes. \textcolor{blue}{Videos are available on the \href{https://jiayi-wu-leo.github.io/real2sam2real/}{project page}.}
    }
    \label{fig:main_03_exp_cam_control}
\end{figure}

\endgroup

\subsection{High-Dynamic Scenes}

\begin{figure}[!htbp]
    \centering
    \includegraphics[width=0.99\linewidth]{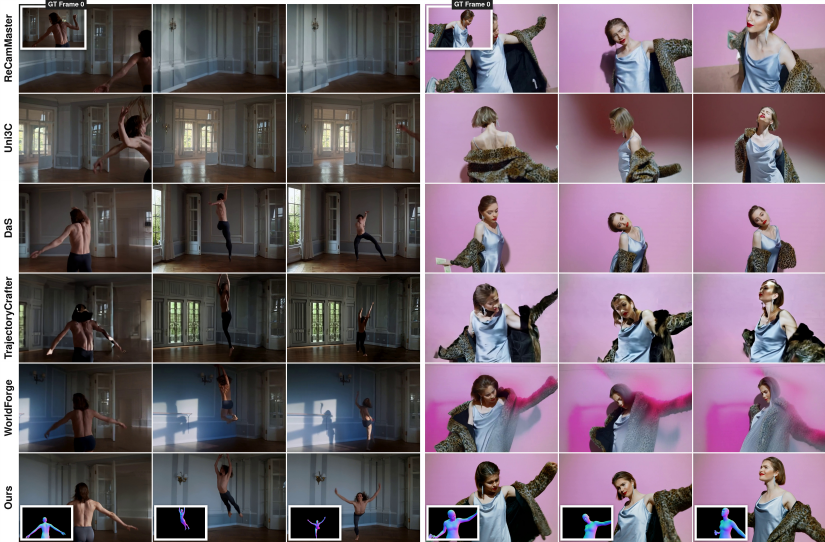}
    \caption{
    \textbf{Qualitative comparison under high-dynamic scene motion.}
    Real2SAM2Real synthesizes coherent video under large non-rigid motion
    coupled with camera motion. This is a challenging regime in which diffusion
    priors alone can break down. All methods reconstruct the real video (last row)
    from the same single input frame (inset, top left); to let every baseline
    operate at its full capability, each is additionally given the strongest
    motion conditioning it supports, i.e. camera poses, point-cloud
    renderings, tracking videos, or a dense warped video. The insets in the
    \textit{Ours} row are the instance-masked normal maps that condition our
    model. The baselines over-rely on diffusion priors and exhibit structural
    collapse and appearance drift, whereas Real2SAM2Real maintains
    spatiotemporal coherence and stays closest to the real video.
    \textcolor{blue}{Videos are available on the \href{https://jiayi-wu-leo.github.io/real2sam2real/}{project page}.}
    }
    \label{fig:main_05}
\end{figure}

We evaluate dynamic scenes, in which non-rigid motion and camera motion occur simultaneously.
With SAM 3D Body~\citep{yang2026sam3dbody}, we obtain a dynamic 3D cache by estimating a per-frame 3D human mesh from the source video, and render its instance-masked normal maps as our condition.
The source video thereby serves as ground truth, and each method is tasked with reconstructing it from a single input frame.
We evaluate on 300 real-world clips from HumanVid~\citep{wang2024humanvid} and OpenVid-HD~\citep{nan2024openvid}, where each baseline additionally receives the strongest motion conditioning it supports, derived from the same clip.
As shown in Tab.~\ref{tab:human_dynamic_quantitative} and Fig.~\ref{fig:main_05}, the baselines exhibit structural collapse and appearance drift under large non-rigid motion, whereas Real2SAM2Real maintains spatiotemporal coherence.
In practice, this setting doubles as a low-cost motion capture pipeline: a casually filmed performance directly yields a dynamic 3D cache, with no dedicated motion-capture equipment.

\providecommand{\first}[1]{\cellcolor{red!12}\textbf{#1}}
\providecommand{\second}[1]{\cellcolor{blue!12}#1}
\begin{table}[!htbp]
\centering
\setlength{\tabcolsep}{3.5pt}
\setlength{\aboverulesep}{0pt}\setlength{\belowrulesep}{0pt}\renewcommand{\arraystretch}{1.05}
\setlength{\abovecaptionskip}{0pt}\setlength{\belowcaptionskip}{3pt}
\caption{Quantitative comparison of deformable human video reconstruction on 300 clips from HumanVid~\citep{wang2024humanvid} and OpenVid-HD~\citep{nan2024openvid}.
Video quality is reported by VBench~\citep{huang2024vbench}, and human pose accuracy (PA-MPJPE~\citep{kocabas2020vibe}, AKD-2D, MKR) by extending the 2D keypoint protocol of Monkey-Net~\citep{siarohin2019animating} to 3D.
Poses are estimated by SAM~3D~Body~\citep{yang2026sam3dbody}.
The best and second-best results are shaded in \colorbox{red!12}{red} and \colorbox{blue!12}{blue}.}
\label{tab:human_dynamic_quantitative}
\resizebox{\textwidth}{!}{%
\begin{tabular}{l cccc ccc}
\toprule
\multirow{2}{*}{\textbf{Method}} & \multicolumn{4}{c}{\textbf{VBench} ($\uparrow$)} & \multicolumn{3}{c}{\textbf{Human Pose Accuracy} ($\downarrow$)} \\
\cmidrule(lr){2-5} \cmidrule(lr){6-8}
 & Subject & Background & Aesthetic & Imaging & PA-MPJPE (mm) & AKD-2D (px) & MKR (\%) \\
\midrule
ReCamMaster~\citep{bai2025recammaster} & 90.47 & 90.80 & 54.41 & 65.81 & \second{34.81} & 78.09 & 9.29 \\
Uni3C~\citep{cao2025uni3c} & 89.07 & 90.21 & 54.12 & 64.05 & 59.45 & 108.58 & 17.85 \\
DaS~\citep{gu2025diffusion} & \first{92.58} & \first{92.07} & \second{55.04} & \second{67.22} & 55.33 & 57.09 & 3.48 \\
TrajectoryCrafter~\citep{yu2025trajectorycrafter} & 90.12 & 90.59 & 52.94 & 63.76 & 37.30 & \second{28.37} & \second{1.91} \\
WorldForge~\citep{song2025worldforge} & 89.81 & 90.80 & 51.91 & 61.74 & 50.59 & 48.19 & 6.12 \\
\midrule
\textbf{Ours} & \second{91.83} & \second{91.12} & \first{56.81} & \first{69.63} & \first{22.80} & \first{17.80} & \first{0.58} \\
\bottomrule
\end{tabular}%
}
\end{table}

\clearpage
\subsection{Entity Motion Control \& Manipulation}

\begin{figure}[!htbp]
    \centering
    \includegraphics[width=0.99\linewidth]{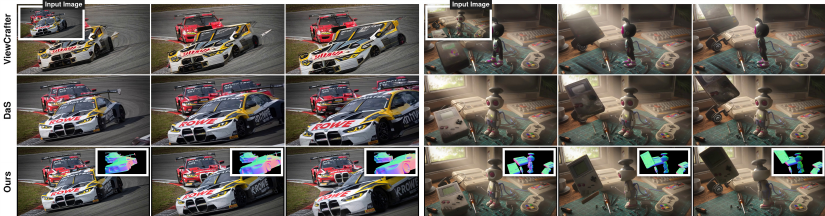}
    \caption{
    \textbf{Qualitative comparison of object manipulation.}
    Two scenes are shown side by side; within each, frames advance in time from left to right. Baseline methods exhibit cascading 
    artifacts and structural breakdowns under large entity movements and 
    occlusions. \textcolor{blue}{Videos are available on the \href{https://jiayi-wu-leo.github.io/real2sam2real/}{project page}.}
    }
    \label{fig:main_04}
\end{figure}

We construct a test set of 100 cases, each specifying randomized 
object translation and rotation trajectories within a GLB scene, with ground-truth 6-DoF poses directly recorded from the environment. We compare against DaS~\citep{gu2025diffusion} and ViewCrafter~\citep{yu2024viewcrafter}.
While ViewCrafter is not designed for entity manipulation, it can perform the task when conditioned on a warped video proxy, reaching pose accuracy comparable to DaS. 
Object pose accuracy is measured by Rotation Error (RotErr) and Translation Error (TransErr) against ground-truth GLB trajectories, with VBench metrics for video quality.
As shown in Tab.~\ref{tab:object_manipulation_quantitative} and Fig.~\ref{fig:main_04}, baseline methods suffer from structural breakdowns and degradation under large entity movements and occlusions, whereas 
Real2SAM2Real maintains precise control and spatiotemporal coherence.

\providecommand{\first}[1]{\cellcolor{red!12}\textbf{#1}}
\providecommand{\second}[1]{\cellcolor{blue!12}#1}
\begin{table}[!htbp]
\centering
\setlength{\tabcolsep}{3.5pt}
\setlength{\aboverulesep}{0pt}\setlength{\belowrulesep}{0pt}\renewcommand{\arraystretch}{1.05}
\setlength{\abovecaptionskip}{0pt}\setlength{\belowcaptionskip}{3pt}
\caption{Quantitative comparison of object manipulation.
Video quality is reported by VBench~\citep{huang2024vbench}, and object 6-DoF accuracy by rotation and translation errors (RotErr, TransErr) of object poses tracked with FoundationPose~\citep{wen2024foundationpose}.
The best and second-best results are shaded in \colorbox{red!12}{red} and \colorbox{blue!12}{blue}; ablations are not ranked.}
\label{tab:object_manipulation_quantitative}
\resizebox{\textwidth}{!}{%
\begin{tabular}{l cccc cc}
\toprule
\multirow{2}{*}{\textbf{Method}} & \multicolumn{4}{c}{\textbf{VBench} ($\uparrow$)} & \multicolumn{2}{c}{\textbf{Object 6-DoF Accuracy} ($\downarrow$)} \\
\cmidrule(lr){2-5} \cmidrule(lr){6-7}
 & Subject & Background & Aesthetic & Imaging & RotErr & TransErr \\
\midrule
DaS~\citep{gu2025diffusion} & \second{96.37} & \second{96.66} & 60.23 & 70.99 & 22.7625 & 48.1813 \\
ViewCrafter~\citep{yu2024viewcrafter} & 94.12 & 94.50 & \second{60.24} & \second{71.32} & \second{21.1082} & \second{41.9168} \\
\midrule
Ours (w/o Soft Injection) & 91.23 & 93.16 & 59.38 & 69.45 & 7.4432 & 17.3950 \\
Ours (w/o Perturbation) & 93.93 & 94.89 & 61.48 & 73.39 & 11.9049 & 22.6497 \\
Ours (w/o Both) & 91.92 & 93.12 & 59.91 & 71.20 & 9.9061 & 21.0134 \\
\midrule
\textbf{Ours (Full)} & \first{96.39} & \first{97.08} & \first{61.21} & \first{73.42} & \first{4.1932} & \first{12.0991} \\
\bottomrule
\end{tabular}%
}
\end{table}

\subsection{Ablation Study}
\begin{figure}[!htbp]
    \centering
    \includegraphics[width=0.99\linewidth]{figures/Ablation.pdf}
    \caption{Ablation on Soft Spatial-Aligned Injection and Instance-Level Spatial Perturbation. \textcolor{blue}{Videos are available on the \href{https://jiayi-wu-leo.github.io/real2sam2real/}{project page}.}}
    \label{fig:Ablation}
\end{figure}

\paragraph{Soft Spatial-Aligned Injection}
To evaluate our geometry injection strategy, we replace the additive geometry branch with channel concatenation (``w/o Soft Injection''). As shown in Fig.~\ref{fig:Ablation}, this variant exhibits appearance distortion and color bleeding near coarse or misaligned normal boundaries. Compared with the full model, it yields lower scores on all four VBench metrics and higher pose errors in both camera and object control (Tables~\ref{tab:camera_control_quantitative} and~\ref{tab:object_manipulation_quantitative}). Our separate geometry branch works alongside the native inpainting pathway, better preserving reference appearance while incorporating 3D guidance.

\paragraph{Instance-level Spatial Perturbation}
The ``w/o Perturbation'' variant removes the instance-level random perturbations applied to normal maps during training. In Fig.~\ref{fig:Ablation}, this variant exhibits appearance distortion around misaligned normal boundaries, indicating greater sensitivity to the gap between training normals and normals rendered from lifted 3D caches. Removing perturbations lowers subject and background consistency and increases pose errors in both tasks (Tables~\ref{tab:camera_control_quantitative} and~\ref{tab:object_manipulation_quantitative}). These results support using perturbed normals to improve robustness to imperfect geometric guidance.

The ``w/o Both'' variant also exhibits appearance distortion (Fig.~\ref{fig:Ablation}). Compared with all three ablation variants, the full model achieves higher subject and background consistency and lower pose errors in both tasks, supporting the combined use of soft injection and normal perturbations.

\subsection{Robotics Application}
\label{sec:robot_app}
\begin{figure}[!htbp]
  \centering
  \includegraphics[width=0.99\textwidth]{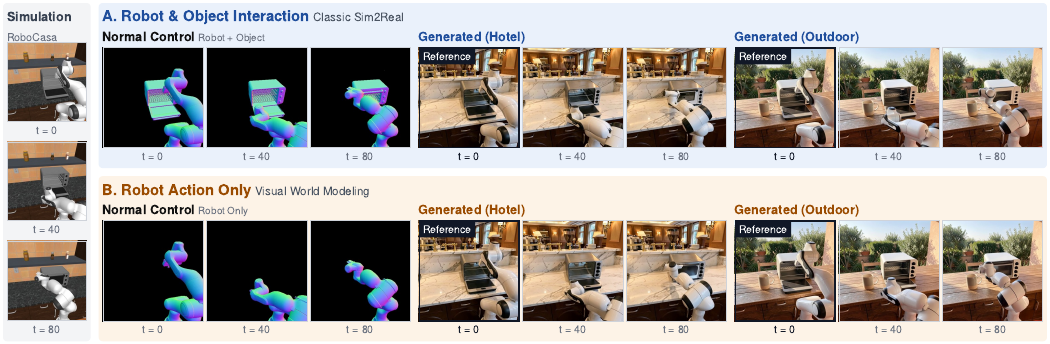}
  \caption{\textbf{Robotics application.}
  Left: a simulated RoboCasa~\citep{robocasa365} episode (``Close the toaster oven door.''), frames $t{=}0,40,80$ of 81.
  Rows: the normal-map control and the videos Real2SAM2Real generates from it in two scenes (hotel coffee station, outdoor terrace) at the same time steps.
  (A) the control contains the robot and the interacted object (Sim2Real); (B) it contains only the robot (visual world modeling).
  The frame tagged \emph{reference} is the input reference image; all other frames are generated. \textcolor{blue}{Videos are available on the \href{https://jiayi-wu-leo.github.io/real2sam2real/}{project page}.}}
  \label{fig:robot_app}
\end{figure}

As a preliminary application, we use Real2SAM2Real for Sim2Real and visual world modeling without robot-specific training (Fig.~\ref{fig:robot_app}). A RoboCasa~\citep{robocasa365} episode is rendered to normal-map control, while FLUX.2-klein-9B generates reference images for different scenes from the corresponding simulated frame while preserving its layout.

For Sim2Real, the control contains both the robot and the interacted object (Fig.~\ref{fig:robot_app}A), preserving the simulated interaction across reference-defined environments. For visual world modeling, only the robot is controlled (Fig.~\ref{fig:robot_app}B), and the model infers the object's response from the reference image and arm motion; the toaster oven door closes consistently with the arm in both scenes. These qualitative examples suggest potential for realistic manipulation video generation and limited visual world modeling; systematic evaluation remains future work.

\section{Conclusion}
We presented Real2SAM2Real, a highly controllable 3D-aware video generation framework that addresses the structural collapse and generative hallucinations of video diffusion models (VDMs) under high-dynamic motion and complex occlusions.
Our coarse 3D foreground cache uses topologically complete meshes for intuitive animation. Normal maps rendered under specified camera trajectories and foreground motions guide the VDM, which synthesizes appearance while largely retaining its pretrained priors.
Adapted from around 2,000 training clips, Real2SAM2Real maintains spatiotemporal coherence across camera control, entity manipulation, and high-dynamic scenes, even under large viewpoint shifts, severe occlusions, and non-Lambertian effects.

\clearpage
\bibliographystyle{preprint}
\bibliography{references}
\clearpage
\appendix
\begin{center}
  {\Large\bfseries Appendix}
\end{center}

\section{Video Results}
Since static images are inherently limited in conveying the spatiotemporal coherence and perceptual quality of generated videos, \textcolor{blue}{we highly encourage readers to visit the \href{https://jiayi-wu-leo.github.io/real2sam2real/}{project page} to view the comprehensive video synthesis results.}

We provide nearly 100 qualitative video comparisons.
All video results corresponding to the experiments in the main paper are available on the \href{https://jiayi-wu-leo.github.io/real2sam2real/}{project page}, covering the following scenarios and applications:
\begin{itemize}
    \item \textbf{Camera Control under Extreme Trajectories:} Camera control along long-range trajectories with large rotations and translations.
    \item \textbf{Camera Control:} Large viewpoint changes with occlusions, compared with the baselines.
    \item \textbf{Reflection, Refraction, and Fluid Dynamics:} Scenes with non-Lambertian surfaces and time-varying effects (e.g., smoke and fire), whose appearance is synthesized by the video diffusion model rather than encoded in the 3D cache.
    \item \textbf{Dynamic Scenes:} Reconstruction of real videos with large non-rigid motion and camera motion from a single frame, compared side by side with all baselines.
    \item \textbf{Motion Transfer:} Editing the reference image while keeping the normal-map sequence fixed to transfer motion across different foreground appearances, e.g., retargeting a captured human performance to a new character.
    \item \textbf{Object Manipulation:} 3D editing of individual foreground entities, such as 6-DoF rigid transformations.
    \item \textbf{Joint Control via 3D Guidance and Text:} Joint control by 3D geometric guidance and text prompts that specify stylistic or environmental effects.
    \item \textbf{Ablation Studies:} Visual comparisons of the effects of soft spatial-aligned injection and instance-level perturbation.
    \item \textbf{Robotics Application:} Videos generated from normal-map controls of robot manipulation episodes (e.g., closing a toaster oven door), illustrating potential applications in Sim2Real and visual world modeling.
\end{itemize}

\section{Implementation Details}\label{supp_impl}

\subsection{Training Data Curation}\label{supp_impl_data}
To construct paired RGB–normal training data, we apply the 3D-free pipeline described in Sec.~\ref{sec3:data_curation} to in-the-wild monocular videos from DL3DV~\citep{ling2024dl3dv}, HumanVid~\citep{wang2024humanvid}, and OpenVid-HD~\citep{nan2024openvid}. Specifically, given a video, objects are identified
by Qwen2-VL~\citep{wang2024qwen2}, and YOLO-World~\citep{cheng2024yolo} localizes
each instance via text-conditioned bounding boxes, from which SAM~3~\citep{carion2025sam} generates
pixel-level instance masks. NormalCrafter~\citep{bin2025normalcrafter} estimates temporally consistent normals, to which we apply the instance masks to form RGB--normal training pairs. In total, we use around 2,000 video clips, with approximately 13\% from DL3DV, 17\% from HumanVid, and 70\% from OpenVid-HD. We uniformly sample 81 frames from each clip.

\subsection{Training Setup}\label{supp_impl_train}
We implement our framework based on Wan2.2-Fun-A14B-InP and fine-tune it on two NVIDIA H200 GPUs (batch size of 1, BF16 mixed precision) for approximately two days. We report results using the checkpoint at training step 6,000. The high- and low-noise experts are fine-tuned separately, each with its own LoRA modules ($r=64$, $\alpha=32$) and geometry patch embedding. We use AdamW with a cosine learning-rate schedule and $\epsilon=10^{-10}$. LoRA uses $lr=10^{-4}$ and weight decay $0.03$, while the geometry embedding uses $lr=10^{-5}$ and no weight decay. Training uses 720p videos of 81 frames, with independent 10\% dropout for the normal, reference-frame, and text conditions.

The flow-matching loss masks extreme velocity residuals: $\rho(e)=D^{-1}\sum_{i=1}^{D} e_i^2\mathbf{1}(|e_i|\leq 50)$, where $D$ is the number of scalar residuals. The mean is taken over all elements, including masked entries.

\section{Experimental Setup}\label{supp_exp}
Each experiment is described in the same order: test data, baselines and their conditioning, and evaluation metrics.

\subsection{Camera Control}\label{supp_exp_cam}
\paragraph{Test Data}
We source 100 in-the-wild images from Unsplash, spanning indoor scenes, outdoor environments, and object-centric compositions.
For each image, we reconstruct the foreground instances as posed meshes with SAM~3D~\citep{chen2025sam} and place them in a GLB scene, with the camera at the source viewpoint.
We sample two camera trajectories per image, yielding 200 test cases.
Each trajectory orbits a pivot on the optical axis at the depth of the foreground meshes; the first trajectory moves in a random in-plane direction, and the second moves horizontally to the left or right.
Each trajectory contains 81 camera poses recorded directly in the GLB scene, so ground-truth poses are available without post-hoc reconstruction.

\paragraph{Baselines and Conditioning}
All methods start from the same source image and follow the same ground-truth trajectory, using their released checkpoints and recommended inference settings.
ReCamMaster~\citep{bai2025recammaster} takes the source image repeated as a static 81-frame video, together with the ground-truth poses converted to its camera format.
ViewCrafter~\citep{yu2024viewcrafter}, TrajectoryCrafter~\citep{yu2025trajectorycrafter}, and WorldForge~\citep{song2025worldforge} share an identical warped video: following the warping procedure of WorldForge, the source image is lifted to a point cloud with the depth and intrinsics estimated during SAM~3D reconstruction and rendered along the ground-truth trajectory.
Each of these methods samples the shared warp at its native length (25, 49, and 49 frames, respectively).
GEN3C~\citep{ren2025gen3c} builds its own 3D cache from the source image and renders it along the ground-truth trajectory at its native length of 121 frames.
Ours renders the instance-masked normal maps of the same GLB scene along the ground-truth trajectory (81 frames).

\paragraph{Evaluation Metrics}
Following~\citep{song2025worldforge, bai2025recammaster}, we measure camera accuracy with Absolute Trajectory Error (ATE) and Relative Pose Error in translation and rotation (RPE-T, RPE-R).
For each generated video, we estimate the camera trajectory from all frames with GLOMAP~\citep{pan2024global} and compare it against the ground truth with evo~\citep{grupp2017evo} after Sim(3) alignment; RPE is computed between consecutive frames.
Video quality is evaluated with VBench~\citep{huang2024vbench}, following~\citep{bai2025recammaster}: we report Subject Consistency, Background Consistency, Aesthetic Quality, and Imaging Quality.

\subsection{High-Dynamic Scenes}\label{supp_exp_dyn}
\paragraph{Test Data}
We collect 300 real-world clips, 100 from HumanVid~\citep{wang2024humanvid} and 200 from OpenVid-HD~\citep{nan2024openvid}, selected for large non-rigid motion coupled with large camera motion.
The source videos are at $1280{\times}720$ and 16 fps, and all methods are evaluated within source frames 0 to 80.
For our method, we obtain the dynamic 3D cache by estimating a per-frame human mesh with SAM~3D~Body~\citep{yang2026sam3dbody}, and render it as instance-masked normal maps.
Since the cache is recovered from the source video, the source video serves as ground truth, and each method reconstructs it from a single frame.

\paragraph{Baselines and Conditioning}
All methods start from a single frame of the source video, and each baseline additionally receives the strongest motion conditioning it supports, derived from the same source video.
ReCamMaster~\citep{bai2025recammaster} receives the camera trajectory estimated from the source video by ViPE~\citep{huang2025vipe}.
Uni3C~\citep{cao2025uni3c} receives the camera trajectory and scene point cloud produced by its own pipeline from the source video.
DaS~\citep{gu2025diffusion} receives the 3D tracking video estimated from the source video.
TrajectoryCrafter~\citep{yu2025trajectorycrafter} and WorldForge~\citep{song2025worldforge} share a dense warped video: using AllTracker~\citep{harley2025alltracker} with frame 0 as the query frame, we track every pixel through the source video and forward-splat the pixels of frame 0 to each subsequent frame, keeping the most confident pixel when several land on the same location and marking uncovered regions as holes.
All pixels of the warped video come from frame 0, so no appearance of the ground-truth video is leaked.
Ours receives the normal-map sequence rendered from the SAM~3D~Body cache.

\paragraph{Temporal and Spatial Alignment}
Since the methods sample frames differently, we compare them only on the source frames covered by all methods, taking the nearest output frame where sampling points differ.
All videos are evaluated on a common grid, obtained by cropping the central $1080{\times}720$ region of the source frame and resizing it to $720{\times}480$.
The native cropping and resizing of each method is determined by matching its output frames against the source video.

\paragraph{Evaluation Metrics}
For human pose accuracy, we run SAM~3D~Body on each generated video and on the ground-truth video with identical crops and box prompts, and compare the two sets of pose estimates.
AKD-2D follows the keypoint protocol of Monkey-Net~\citep{siarohin2019animating} and measures the average 2D distance of 23 body and foot joints on the $720{\times}480$ grid.
PA-MPJPE~\citep{kocabas2020vibe} extends this protocol to 3D: it is computed over the same 23 joints after root removal and similarity (Procrustes) alignment, in millimeters.
MKR is the fraction of frames in which the main subject cannot be detected without a box prompt.
Video quality is evaluated with VBench~\citep{huang2024vbench}: we report Subject Consistency, Background Consistency, Aesthetic Quality, and Imaging Quality.

\subsection{Entity Motion Control}\label{supp_exp_obj}
\paragraph{Test Data}
We construct 100 test cases, each assigning randomized 6-DoF trajectories (translation and rotation) to one or more foreground objects in a GLB scene, while the camera remains static.
Each trajectory spans 81 frames, and ground-truth object poses are recorded directly in the GLB scene without post-hoc reconstruction.

\paragraph{Baselines and Conditioning}
All methods start from the same source image and follow the same object trajectories.
DaS~\citep{gu2025diffusion} receives a 3D tracking video: we lift the source image to a point cloud with monocular geometry estimation, move the points of the manipulated objects along the ground-truth trajectories with the camera fixed, and render the result.
ViewCrafter~\citep{yu2024viewcrafter} is not designed for entity manipulation; we instead provide a warped video proxy, obtained by forward-warping the source image with the same moving point cloud, which enables it to perform the task.
Ours moves the SAM~3D meshes of the objects along the same trajectories and renders their instance-masked normal maps.

\paragraph{Evaluation Metrics}
We track the 6-DoF pose of each manipulated object in the generated videos with FoundationPose~\citep{wen2024foundationpose}, using the SAM~3D mesh of the object as its model; the object is registered in the first frame with its instance mask and tracked in subsequent frames.
Since generated videos have no depth, we estimate per-frame relative depth with Depth Anything V2~\citep{yang2024depth} and calibrate its scale and shift on the first frame against the depth of the mesh rendered at the ground-truth pose, yielding metric depth.
All methods share the camera intrinsics estimated during SAM~3D reconstruction, which are also used to render the ground truth.
Tracked poses are converted to motion relative to the first frame and compared with the ground-truth relative motion, excluding the first frame, so that registration errors in the first frame cancel out and the metrics reflect whether the objects move correctly.

\section{Robustness to 3D Cache Errors}\label{supp_robust}
Single-image 3D reconstruction is inherently under-constrained, and the 3D cache produced by SAM~3D may contain geometric and placement errors.
Our design treats the cache as a coarse spatial anchor rather than an exact rendering constraint: it enters through soft spatial-aligned injection, and training-time instance-level perturbation teaches the model to tolerate misaligned normals (Sec.~\ref{sec3:data_curation}).
To quantify this tolerance, we inject controlled errors into the estimated cache on the camera-control benchmark.

\paragraph{Setup}
Each object in the cache is displaced in a random direction by a magnitude $\delta$, given as a percentage of its own bounding-box diagonal.
We evaluate eight levels, $\delta \in \{3, 6, 9, 12, 16, 19, 22, 25\}\%$, on a subset of 30 test cases, and group them into three bands in Tab.~\ref{tab:supp_robust}; all methods, including the baselines, are re-evaluated on this subset, so the numbers differ from those in Tab.~\ref{tab:camera_control_quantitative}.
At the strongest level, objects may float above their supporting surfaces and intersect their neighbors.
Besides camera accuracy and VBench, we report PSNR, SSIM, and LPIPS of the first generated frame against the source image, which reveal whether cache errors propagate into the generated pixels.

\paragraph{Results}
Video quality is largely insensitive to cache errors: VBench stays at 0.84, and first-frame PSNR, SSIM, and LPIPS remain flat, although the displacement grows about fivefold from the lowest to the highest band.
Camera accuracy degrades gradually: ATE rises monotonically from 0.015 to 0.027, indicating that the cache does anchor the spatial layout, while RPE-T and RPE-R remain nearly unchanged.
Even at the strongest corruption, our ATE and RPE-T remain lower than those of all warp-based baselines evaluated on the same cases.

\begin{table}[!htbp]
\centering
\caption{Robustness to 3D cache errors on 30 camera-control test cases. $\delta$: random displacement of each cached object, as a percentage of its bounding-box diagonal. PSNR, SSIM, and LPIPS compare the first generated frame with the source image. Since all methods are evaluated on this subset, the numbers are not directly comparable to Tab.~\ref{tab:camera_control_quantitative}.}
\label{tab:supp_robust}
\resizebox{\textwidth}{!}{%
\begin{tabular}{l c ccc c ccc}
\toprule
\textbf{Method} & $\delta$ (\%) & PSNR $\uparrow$ & SSIM $\uparrow$ & LPIPS $\downarrow$ & VBench $\uparrow$ & ATE $\downarrow$ & RPE-T $\downarrow$ & RPE-R $\downarrow$ \\
\midrule
Ours & 0      & 30.63 & 0.8998 & 0.0669 & 0.838 & 0.015 & 0.014 & 1.087 \\
Ours & 3--6   & 30.71 & 0.8998 & 0.0661 & 0.840 & 0.017 & 0.014 & 1.066 \\
Ours & 9--16  & 30.69 & 0.8998 & 0.0660 & 0.840 & 0.023 & 0.015 & 1.058 \\
Ours & 19--25 & 30.61 & 0.8994 & 0.0662 & 0.840 & 0.027 & 0.015 & 1.038 \\
\midrule
TrajectoryCrafter~\citep{yu2025trajectorycrafter} & -- & -- & -- & -- & 0.822 & 0.034 & 0.029 & 1.811 \\
ViewCrafter~\citep{yu2024viewcrafter}             & -- & -- & -- & -- & 0.829 & 0.054 & 0.062 & 3.756 \\
WorldForge~\citep{song2025worldforge}             & -- & -- & -- & -- & 0.816 & 0.076 & 0.039 & 1.920 \\
\bottomrule
\end{tabular}%
}
\end{table}

\section{Limitation}\label{limit}
Our framework supports human motion and deformable objects, but the assets produced by most 3D lifting models are not directly simulation-ready. Accurately simulating complex interactions between objects, contact-induced deformation, or fracture remains challenging. In these settings, a coarse 3D cache may provide misleading geometric guidance, limiting the physical plausibility and controllability of the generated interactions.

Appearance is directly anchored only at the reference frame. For other frames, the VDM must propagate reference appearance using its spatiotemporal priors and coarse geometric guidance, without an equally direct appearance constraint. Fine textures and instance identity may therefore drift under large viewpoint changes or prolonged occlusions, while the appearance of newly revealed surfaces depends on generative inference.

\end{document}